\documentclass[sigconf]{aamas}

\usepackage{balance} 

\usepackage{algorithm}
\usepackage{caption}
\usepackage{subcaption}
\usepackage{float}
\usepackage{algpseudocode}
\usepackage{amsmath}

\usepackage{appendix}

\DeclareFontFamily{U}{stix2bb}{}
\DeclareFontShape{U}{stix2bb}{m}{n} {<-> stix2-mathbb}{}

\NewDocumentCommand{\indicator}{}{\text{\usefont{U}{stix2bb}{m}{n}1}}

\DeclareMathOperator*{\argmax}{arg\,max} 
\newcommand{\etal}{et al. }

\makeatletter
\gdef\@copyrightpermission{
  \begin{minipage}{0.2\columnwidth}
   \href{https://creativecommons.org/licenses/by/4.0/}{\includegraphics[width=0.90\textwidth]{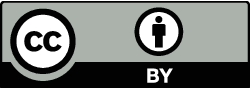}}
  \end{minipage}\hfill
  \begin{minipage}{0.8\columnwidth}
   \href{https://creativecommons.org/licenses/by/4.0/}{This work is licensed under a Creative Commons Attribution International 4.0 License.}
  \end{minipage}
  \vspace{5pt}
}
\makeatother

\setcopyright{ifaamas}
\acmConference[AAMAS '25]{Proc.\@ of the 24th International Conference
on Autonomous Agents and Multiagent Systems (AAMAS 2025)}{May 19 -- 23, 2025}
{Detroit, Michigan, USA}{Y.~Vorobeychik, S.~Das, A.~Nowe (eds.)}
\copyrightyear{2025}
\acmYear{2025}
\acmDOI{}
\acmPrice{}
\acmISBN{}

\acmSubmissionID{1265}

\title[AAMAS-2025 Formatting Instructions]{Conformal Set-based Human-AI Complementarity with Multiple Experts}

\author{\href{https://orcid.org/0000-0002-9439-1262}{Helbert Paat}}
\affiliation{
  \institution{The Hong Kong Polytechnic University}
  \city{Hong Kong SAR}
  \country{China}}
\email{hapaat@polyu.edu.hk}

\author{\href{https://orcid.org/0000-0002-5277-0176}{Guohao Shen}}
\affiliation{
  \institution{The Hong Kong Polytechnic University}
  \city{Hong Kong SAR}
  \country{China}}
\email{guohao.shen@polyu.edu.hk}

\begin{abstract}
Decision support systems are designed to assist human experts in classification tasks by providing conformal prediction sets derived from a pre-trained model. This human-AI collaboration has demonstrated enhanced classification performance compared to using either the model or the expert independently. In this study, we focus on the selection of instance-specific experts from a pool of multiple human experts, contrasting it with existing research that typically focuses on single-expert scenarios. We characterize the conditions under which multiple experts can benefit from the conformal sets. With the insight that only certain experts may be relevant for each instance, we explore the problem of subset selection and introduce a greedy algorithm that utilizes conformal sets to identify the subset of expert predictions that will be used in classifying an instance. This approach is shown to yield better performance compared to naive methods for human subset selection. Based on real expert predictions from the CIFAR-10H and ImageNet-16H datasets, our simulation study indicates that our proposed greedy algorithm achieves near-optimal subsets, resulting in improved classification performance among multiple experts.

\end{abstract}

\keywords{Prediction Sets; Conformal Prediction Sets; Human-AI Team; Multiple Experts; Human-AI Interaction; Confusion Matrix; Multiclass Classification; Subset Selection}

\newcommand{\BibTeX}{\rm B\kern-.05em{\sc i\kern-.025em b}\kern-.08em\TeX}

\begin{document}


\pagestyle{fancy}
\fancyhead{}


\maketitle 


\section{Introduction}
\label{1.intro}

In recent years, human experts have increasingly relied on AI-based decision support systems to make informed choices in high-risk fields such as medicine, drug discovery, finance, law, and science \cite{Jiao2020ADL, Liu2021EvaluatingEC, GrgicHlaca2019HumanDM, llm_aamas, davide_aamas24}. Although much existing research focuses on developing sophisticated algorithms, it is essential to advance the paradigm of AI-assisted decision making, where humans and AI collaborate to improve accuracy. Given the impressive performance of modern machine learning models, it is crucial for humans to learn how to effectively leverage these tools for important real-world tasks. This collaboration, known as \textit{human-AI complementarity}, can lead to better outcomes than when humans and machines operate independently.

Previous studies aim to achieve human-AI complementarity by suggesting that decision support systems should help individuals identify situations where AI offers substantial advantages and provide explanations that clarify the reasoning behind model predictions \cite{Wang2021AreEH, Vodrahalli2022UncalibratedMC, Papenmeier2019HowMA, subbarao_aamas19, ibrahim_expla_aamas23}. This assistance often involves examining the factors that influence trust \cite{Suresh2020MisplacedTM, Zhang2020EffectOC, Yin2019UnderstandingTE, siddharth_aamas21_trust}. However, these investigations have yielded inconclusive results, leading to uncertainty about how experts can mitigate the risk of developing misplaced trust in AI systems.
A study conducted by \citet{Straitouri2022ImprovingEP} developed a system designed to operate without requiring experts to discern when and how to trust AI, thus shifting the focus of the human-AI model from factors such as calibration and explanation. This system generates a set of label predictions, referred to as a prediction set, from which a human expert selects the most appropriate label. In their proposed framework, the construction of these prediction sets is based on conformal predictors \cite{Angelopoulos2023ConformalPA}. Furthermore, \citet{Toni2024TowardsHC} suggest that the conformal predictor may not be the optimal set-valued predictor in this kind of system. They propose a framework for constructing optimal prediction sets that enable human experts to achieve the highest possible accuracy. However, both of these studies are limited in scope, focusing solely on a single expert. We argue that this limitation restricts the potential accuracy and practicality of combined human-AI models. In practice, decision-making typically comes from multiple experts who engage with decision support systems.
Our goal is to establish a framework for human-AI complementarity that involves multiple human experts working with a decision support system. For each instance, we propose a human subset selection algorithm and design a framework that incorporates predictions from this subset of multiple experts to generate final predictions in multiclass classification tasks, thereby enhancing decision-making outcomes in complex scenarios.

\begin{figure*}[t!]
  \centering
  \includegraphics[width=18cm]{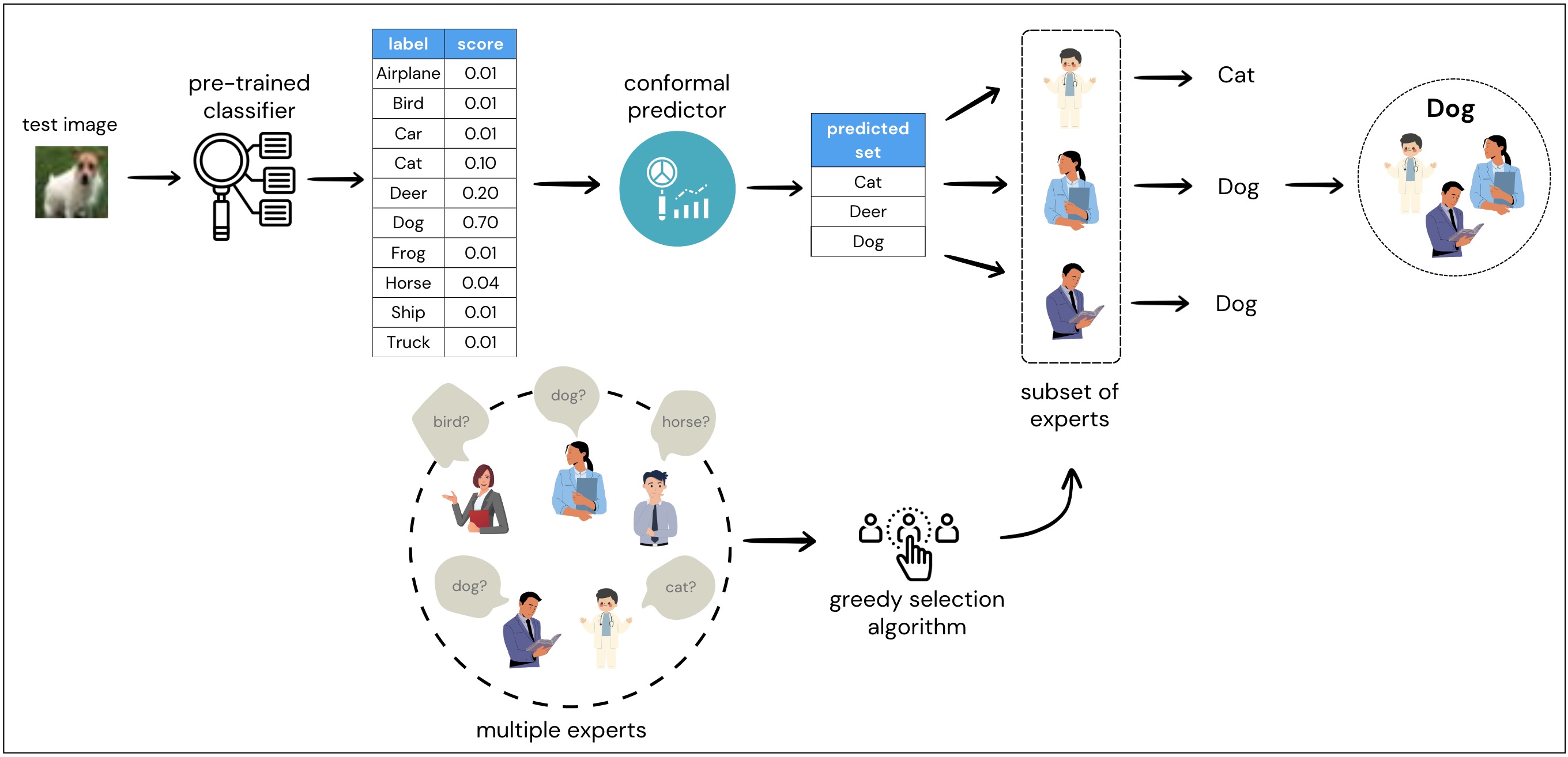}
  \caption{
Illustration of the proposed greedy subset selection algorithm that operates within a multiple expert framework utilizing conformal sets during inference. Initially, a pre-trained classifier computes scores for the conformal predictor. For a given test image, the greedy selection algorithm identifies a subset of human experts. Each selected expert then makes their decision from a narrowed set of options. The final prediction is determined through a combination policy, typically employing a majority decision rule to merge the predictions from multiple experts.}
\Description{Illustration/visualization of the proposed greedy subset selection algorithm showing the conformal predictor narrowing the option and the proposed subset selection algorithm choosing subset of experts.}
\label{fig:framework}
\end{figure*}

\textbf{Our contributions}. We explore the scenario of multiple human experts collaborating with a decision support system that offers a set of label predictions and requires each expert to select from this set during inference. We begin by establishing a lower bound on the accuracy of a system that incorporates the experts' predictions that are chosen from the conformal sets. Then we identify the conditions that allow these experts to effectively utilize conformal sets instead of relying on the entire label space. We then proceed with the perspective of subset selection of human experts, asking which subset of experts should be chosen to classify each data sample. Inspired by these findings, we propose a greedy algorithm for selecting a subset of human experts for each instance, leveraging the conformal prediction set to attain near-optimal classification performance. Intuitively, the goal is to favor human experts who, even without prior knowledge of the conformal set, are more inclined to select elements from within that set. This, however, assumes that we understand the likelihood of each expert choosing specific classes based on the true class\footnote{We achieve this through the confusion matrix for an expert, which illustrates how an instance is categorized by human experts into different classes given the true label. The conformal set narrows the possible labels for a human, and the confusion matrix is important to assess the relative importance of each label within the conformal set, excluding irrelevant labels.}.

Finally, we validate the effectiveness of our approach through simulation studies using real expert predictions on the CIFAR-10H and Imagenet-16H datasets on several multiclass classification tasks. We demonstrate that, for varying numbers of human experts, our proposed greedy selection algorithm consistently outperforms naive approaches of choosing human subsets, such as selecting random subset of humans or choosing the subsets that favor the majority's prediction. This suggests that a strategic human expert selection process can improve classification accuracy. Moreover, we also demonstrate that our approach outperforms subset selection based on top-$k$ set predictors where the top-$k$ sets are the basis for the selection of human subsets. We also show that even as we increase the number of experts, the proposed algorithm is still useful for selecting the appropriate human experts for each instance. Figure \ref{fig:framework} shows the general illustration of the proposed framework.

\section{Related Works}
\label{2.related_work}

In literature, we have seen the emergence of methods that both utilize AI and human expert decisions in prediction tasks to leverage their respective strengths -- one such area is the development of classifiers that perform predictions to some samples and rely on the human experts for the remaining ones through a triage policy \cite{Raghu2019TheAA, Mozannar2020ConsistentEF, Gao2021HumanAICW, Verma2022CalibratedLT, De2020ClassificationUH, Okati2021DifferentiableLU, Charusaie2022SampleEL, Bansal2021IsTM}. One common approach involves learning a binary classifier that defers instances with low model confidence to humans \cite{Cortes2016LearningWR, Hendrycks2016ABF}. A class of algorithms known as "learning to defer" (L2D) focuses on training models that adapt to human experts by learning when to either make a prediction themselves or defer the decision to the experts. L2D approaches have been extensively researched \cite{Mozannar2023WhoSP, Dvijotham2023EnhancingTR, Hemmer2023LearningTD} and have been adapted for scenarios involving multiple human experts \cite{Mao2023TwoStageLT, Hemmer2022FormingEH, Mao2023PrincipledAF, Keswani2021TowardsUA}. However, a notable limitation of this body of work is that AI models tend to excel only in handling instances with high confidence, which restricts their generalization capabilities across all instances. More relevant to our study, \citet{Babbar2022OnTU} propose decision support systems that utilize prediction sets. Unlike their approach, however, our pre-trained classifier assists human experts in solving classification tasks by suggesting prediction sets from which the selected experts can select labels for each data sample during inference.

There is a substantial body of research focused on \textit{set-valued predictors}, which aim to develop models that produce a set of label values known as a prediction set \cite{Chzhen2021SetvaluedC}. In several studies on cautious or reliable classification \cite{Mortier2019EfficientSP, Nguyen2021MultilabelCW, Ma2021PartialCI}, the model-generated set-valued predictions provide a way to understand the uncertainty in the model's predictions. Recent works have explored decision support systems that utilize prediction sets to assist human experts \cite{Straitouri2022ImprovingEP, Babbar2022OnTU, Cresswell2024ConformalPS, Zhang2024EvaluatingTU, Toni2024TowardsHC}, demonstrating that set-valued predictors can enhance the performance of human experts in prediction tasks. Notably, conformal predictors—a type of set-valued predictor with distribution-free guarantees—have shown promise in this context. While the studies by \cite{Straitouri2022ImprovingEP, Toni2024TowardsHC} are closely related to our work, we specifically focus on scenarios involving multiple experts, contrasting their single-expert setting.

Another line of research \cite{Kerrigan2021CombiningHP, Singh2023OnSS} enhances classification performance by directly integrating human predictions with the output probabilities of a pre-trained classifier during inference. In this framework, data samples are not deferred to either the human or the model; instead, predictions are made through a mathematical combination of the human label and the classifier's probabilities. \citet{Kerrigan2021CombiningHP} combine the probabilistic outputs of an AI model with human class-level outputs to improve the accuracy of the human-AI collaboration. More closely related to our work, \citet{Singh2023OnSS} propose a greedy algorithm that incorporates class-level outputs from multiple human experts alongside the probabilistic outputs of a pre-trained classifier. Their approach identifies the optimal subset of workers for a task, enhancing the combined human-AI decision model. Similarly, we also view multiple expert collaboration as a subset selection problem; however, our focus is on selecting human subsets that will perform classification based on knowledge of conformal prediction sets.
\section{Preliminary}

\subsection{Problem Formulation}

In our setting for human-AI multiclass classification tasks, a set of $h$ human experts $\mathcal{H}$ aims to predict the corresponding label $Y \in \mathcal{Y} = \{1,...,n\}$ from the feature vector $X \in \mathcal{X}$, given access to an automated decision support system $\mathcal{C}:\mathcal{X} \rightarrow 2^\mathcal{Y}$ that predicts a set of potential labels $\mathcal{C}(X) \subseteq\mathcal{Y}$. Here, human experts' predictions for the label of $X$ are $H(X)=\{H_i(X)\}_{i=1}^{h}$ where $H_i(X)\in\mathcal{Y} = \{1,...,n\}$ and the system $\mathcal{C}$ is usually a conformal predictor \citep{Angelopoulos2021AGI, Vovk2005AlgorithmicLI}, which helps human experts by narrowing the scope of prediction. The system $\mathcal{C}$  established based on a pre-trained classifier $\hat{f}: \mathcal{X}\rightarrow[0,1]^{|\mathcal{Y}|}$, where $\hat{f}(X)$ is the normalized probability vector of the prediction on $X$ and $\mathcal{C}(X)$ is determined based on the outputs scores $\hat{f}(X)$. Given the system $\mathcal{C}$, our objective is to develop a framework to select a subset $\mathcal{S}(X)$ from human experts $\mathcal{H}$.



Similar to \citet{Straitouri2022ImprovingEP}, the system requires that the final prediction of each expert $H_i(X) \in \mathcal{Y}$ be an element of the narrowed set $\mathcal{C}(X).$ Multiple human predictions $\{H_i(X)\}_{i\in\mathcal{S}(X)}$ form the basis for generating the final experts prediction $\hat{Y}$.

Ideally, we expect the designed framework to benefit from the collaborative predictions of experts in subset $\mathcal{S}(X)$ in such a way that:

\begin{center}
\small
$\mathbb{P}[\hat{Y}=Y;\mathcal{C}|H, Y \in \mathcal{C}(X)] \geq \mathbb{P}[H_i=Y;\mathcal{C}|Y \in \mathcal{C}(X)] \geq \mathbb{P}[H_i=Y; \mathcal{Y}]$ 
\end{center}
for any $i \in \{1,...,h\}$ where $\mathbb{P}[\hat{Y}=Y;\mathcal{C}|H, Y \in \mathcal{C}(X)]$ indicates the success probability of the subset $\mathcal{S}(X)$ of multiple experts who predicts a class from the narrowed subset $\mathcal{C}(X)$, and $\mathbb{P}[H_i=Y;\mathcal{C}|Y \in \mathcal{C}(X)]$ indicates the success probability of the $i$th single expert who chooses from the narrowed options $\mathcal{C}(X)$. Subset selection of human experts arises from the intuition that within a team of experts, diverse perspectives may exist. Hence, it becomes necessary to choose a specific subset $\mathcal{S}(X)$ for classification in any given instance. One might instinctively favor selecting the subset $\mathcal{S}(X) = \mathcal{H}$ and employing a majority decision rule to get a final prediction. However, this approach may not be the most effective option, as we will show later.

\subsection{Conformal Prediction}

Given a calibration dataset $D_{\text{cal}} = (x_i, y_i)_{i=1}^l$ and a test sample $(x_{test}, y_{test})$, conformal prediction aims to construct a prediction set $\mathcal{C}$ based on $D_{\text{cal}}$ such that the marginal coverage at a user-specified tolerance level $\alpha \in [0,1]$ is satisfied, i.e.

\begin{equation}
\mathbb{P}[y_{test} \in \mathcal{C} (x_{test})] \geq 1- \alpha.
\end{equation}

Constructing a conformal predictor usually requires calculating the conformal score $s:\mathcal{X} \times \mathcal{Y} \rightarrow \mathbb{R}$ by $s(x_i, y_i) = 1 - \hat{f}_{y_i}(x_i)$ on each sample, where $\hat{f}$ denotes a pre-trained classifier and $\hat{f}_{y_i}(x_i)$ denotes the predicted probability for label $y_i$ given $x_i$. A lower conformal score $s(x_i, y_i)$ means better agreement between the input $x_i$ and the label $y_i$. While a conformal score is close to one implies the pre-trained classifier is significantly incorrect on $x_i$. Let $\hat{q}_{\alpha}$ be the empirical $\lceil (l+1)(1-\alpha) \rceil$-th quantile of the conformal scores $s(x_1, y_1), \ldots, s(x_l, y_l)$. Then the conformal prediction set $C(x_{test})$ is constructed by $C(x_{test}) = \{y : s(x_{test}, y) \leq \hat{q}_\alpha\}$. Note that the constructed set $C(x_{test})$ can be established based on any pre-trained classifier $\hat{f}$ without relying on any distributional assumptions about the data \citep{Vovk2005AlgorithmicLI}. One can show that, conditioned on the calibration set $\mathcal{D}_{\text{cal}}$, the probability of the true label $y_{test}$ belonging to the subset $\mathcal{C}(x_{test})$ is guaranteed at level $(1-\alpha)$ \cite{Angelopoulos2021AGI}.
\section{Utilizing Conformal Prediction Sets for the Selection of Human Subsets}
\label{3_subset_selection}

The multiple experts subset selection problem mathematically involves identifying a specific subset of human experts from a larger set. Let $\mathcal{D}_{test} = \{(x_i, y_i)\}_{i=1}^t$ denote the testing set. Given any test sample $(x,y)$ in the test set $\mathcal{D}_{test}$, the goal is to find a subset $\mathcal{S}(x) \subseteq \mathcal{H}$ with final expert predictions $\{H_i(x)\}_{i=1,\ldots,|\mathcal{S}(x)|}$ that maximizes the experts' conditional success probability $\mathbb{P}[\hat{Y} = y; \mathcal{C} | H, y \in \mathcal{C}(x)]$ where $\hat{Y}=\pi(\{H_i(x)\}_{i=1,\ldots,|\mathcal{S}(x)|})$ is the combination of the selected experts ' prediction based on a given combination method or decision rule $\pi$. This can be expressed as follows:

\begin{center}
   $\max\limits_{\mathcal{S}(x)  \subseteq \mathcal{H} } \: \mathbb{P}[\pi(\{H_i(x)\}_{i=1,\ldots,|\mathcal{S}(x)|}) = y; \mathcal{C} | H, y \in \mathcal{C}(x)]$.
\end{center}

Note that several naive selection methods for forming subsets exist in this case. We can choose the subset $\mathcal{S}(x)$  to be a random subset of $\mathcal{H}$ of fixed size $\gamma < h$ where $h$ is the size of $\mathcal{H}$. The subset $\mathcal{S}(x)$ can also be chosen to be of size $1$, containing only the best-performing expert in the team $\mathcal{H}$. Furthermore, the subset $\mathcal{S}(x)$ can be chosen as the whole expert team $\mathcal{H}$. However, as we will show in the simulation study, these naive approaches to select subsets of humans are not guaranteed to perform the best during inference.

In what follows, our goal is to develop a framework to select subsets of experts using the knowledge of the conformal set and show, theoretically and empirically, that it yields better performance than naive approaches. We first discuss a formulation of the expert’s conditional success probability in the context of multiple expert predictions. To establish the theoretical results, we start by considering the entire team of experts. For a feature vector $x$, let $\{p_i\}_{i=1}^{h}$ be an observation or instantiations of the random variables $H(x)=\{H_i(x)\}_{i=1}^{h}$. In combining multiple human experts' predictions given the conformal set, we assume independence between human predictions. Using the Bayes rule, we can get the following relationship: 

\begin{align}
\nonumber \mathbb{P}[\hat{Y} = y|H, y \in \mathcal{C}(x)] &\propto \mathbb{P}[\hat{Y}=y| y \in \mathcal{C}(x)] \:\:\: \times \\
& \:\:\: \prod\limits_{i \in [h]} \mathbb{P}[H_i = y|\hat{Y}, y \in \mathcal{C}(x)].
\end{align}


Similar to \citet{Straitouri2022ImprovingEP}, the expert’s success probability $\mathbb{P} [\hat{Y} = y | H, y \in \mathcal{C}(x)]$ is estimated by utilizing the multinomial logit model. For a given sample $(x,y)$ and subset $\mathcal{C}(x)$, we assume that the expert’s conditional success probability can be estimated as follows:

\begin{equation}
\mathbb{P}[\hat{Y} = y | H , y \in \mathcal{C}(x)] 
= \prod_{i \in [h]} \mathscr{C}^i_{yp_i}  
= \prod_{i \in [h]} \frac{e^{\mu^i_{yp_i}}}{\sum_{y' \in \mathcal{C}(x)} e^{\mu^i_{y'p_i}}}
\label{eq:multinomial_logit}
\end{equation}
where $ \mu_{yp_i}^i $ denotes the preference of the expert $ i $ for the label value $ p_i \in \mathcal{Y} $, given that the true label is $ y $. We set the parameters $\mu_{yp_i}^i = \log C_{yp_i}^i$ where we assume access to a confusion matrix $C$ of the expert predictions in the multiclass classification task. The confusion matrix is estimated based on real expert predictions using maximum likelihood estimation. More specifically, $C = [C_{yy'}]_{y,y' \in \mathcal{Y}}$, where $C_{yy'} = \mathbb{P}[\hat{Y}=y'; \mathcal{Y} | Y=y]$.
\vspace{2pt}

To determine the improvement in the classification performance of the framework when we have multiple human expert predictions and knowledge of the conformal sets, we derive a lower bound on the accuracy of the framework (see the proof in Appendix \ref{appendix_proofs_1}).
\vspace{2pt}

LEMMA 1. Given $h$ human expert predictions $H=[p_1, p_2, \ldots, p_h]$, conformal set $\mathcal{C}$ with tolerance level $\alpha$, and the $i$th expert’s conditional success probability $\mathscr{C}^i$, the lower bound on the accuracy of the combined framework is given as

\begin{equation}
\mathbb{E}[ \indicator( \{\hat{Y}=y  \} \: \cap \: \{y \in \mathcal{C}\})] \geq \mathbb{P} [ \: \prod\limits _{i \in [h]} \dfrac{\mathscr{C}_{y p_i }^{i}}{1-\mathscr{C}_{y p_i}^{i}} > 1 ] \cdot (1-\alpha).
\label{eq:expectation_bound}
\end{equation}

In the following theoretical result, we demonstrate that, in the approach that uses predictions from multiple experts, utilizing a conformal prediction set $ \mathcal{C} $ within the combination framework results in a tighter lower bound than that derived from considering the entire label space $ \mathcal{Y} $ (as established in Lemma 4.1 of \citet{Singh2023OnSS}), given certain assumptions (see the proof in Appendix \ref{appendix_proofs_2}).\\

LEMMA 2. Let $\epsilon = \prod\limits _{i \in [h]} \dfrac{\mathscr{C}_{yp_i}^{i}}{1-\mathscr{C}_{yp_i}^{i}} - \prod\limits _{i \in [h]} \dfrac{C_{yp_i}^{i}}{1-C_{yp_i }^{i}}$. Assuming $\epsilon > 1$ and $\alpha=0$, the combination framework of multiple human expert predictions that utilizes a conformal set achieves a tighter lower bound on accuracy,

\begin{equation}
\mathbb{P} [ \prod\limits _{i \in [h]} \dfrac{C_{yp_i}^{i}}{1-C_{y p_i}^{i}} > \dfrac{1-\hat{f}_{y}}{\hat{f}_{y}} ] \leq \mathbb{P} [ \: \prod\limits _{i \in [h]} \dfrac{\mathscr{C}_{yp_i}^{i}}{1-\mathscr{C}_{yp_i}^{i}} > 1 ].
\end{equation}








\textbf{Designing a Subset Selection Framework}. Lemma 2 demonstrates the advantage of incorporating the conformal sets $\mathcal{C}$ into the combination framework of multiple expert predictions. Motivated by these theoretical results, we describe a framework for selecting a subset of humans $\mathcal{S}(x)$ instead of all humans $\mathcal{H}$ for each test sample. For the task of human subset selection, let $\hat{p}_i$\footnote{We note that for the subset selection of humans, all experts make initial predictions which are not necessarily in the conformal set. The intuition is to utilize the conformal prediction sets from the decision support system $\mathcal{C}$ for additional guidance in selecting the appropriate human experts. However, during inference, the selected human experts choose their final prediction from the conformal set.} denote the initial prediction of the expert $i$. Following the same intuition as Algorithm 1 in \citet{Singh2023OnSS}, the goal is to select a subset such that the derived lower bound (in Equation \ref{eq:expectation_bound}) is maximized. We maximize the left probability term in the lower bound of Lemma 1. Consider that the term $\mathscr{C}_{y \hat{p}_i }^{i} / (1-\mathscr{C}_{y \hat{p}_i}^{i})$ is greater than $1$ if and only if $\mathscr{C}^i_{y \hat{p}_i} > 0.50$. Hence, to maximize the product of these terms, we choose only the corresponding human predictions that satisfy this constraint. However, the ground truth label $y$ is obviously required to maximize this term in Lemma 1. Since we do not have access to $y$ during test, we choose a pseudo label $y^*$ and define it to be the class that maximizes the probability in the derived lower bound (Equation \ref{eq:expectation_bound}). Although similar in intuition as in \citet{Singh2023OnSS}, we note that, instead of choosing from the full class of labels $\mathcal{Y}$, we choose the pseudo label from the conformal set $\mathcal{C}(x)$. Thus, given the pseudo label $y^*$, we select the pseudo-optimal human subset as follows:

\begin{equation}
\mathcal{S}(x)^* = \argmax\limits_{\mathcal{S}(x)} \:  (\prod\limits_{i \in \mathcal{S}(x)} \dfrac{\mathscr{C}_{y^*\hat{p}_i }^{i}}{1-\mathscr{C}_{y^* \hat{p}_i}^{i}}). 
\end{equation}

\noindent We emphasize that the pseudo label $y^*$ is in the conformal set, that is, $y^* \in \mathcal{C}(x)$. Moreover, we only consider the human predicted labels $\hat{p}_i$'s that are in the conformal set. In addition to excluding humans with corresponding values $\mathscr{C}^i_{y \hat{p}_i}$ of at most $0.50$, knowledge of the conformal set also eliminates any human whose initial predicted label is not in the set.

In Algorithm \ref{alg:greedy_selection}, we present the overall greedy humans subset selection method\footnote{The proposed algorithm is a modified version of the one introduced by \citet{Singh2023OnSS}, where we focus on the set $ \mathcal{C}(x) $ rather than the entire label space $ \mathcal{Y} $. In our simulation study, we will also present results using the top-$k$ sets generated by $ \hat{f} $ instead of the conformal sets $ \mathcal{C}(x) $ for human subset selection.} based on the conformal set. Assume that the worst-case (or maximum) size of any conformal set is $c$. Moreover, we assume that the algorithm has access to  $\mathscr{C}$, which is calculated based on the confusion matrix $C$ by considering only the probabilities of the classes present in  $\mathcal{C}(x)$  and normalizing these probabilities row-wise, as in Eq. \ref{eq:multinomial_logit}. The algorithm then calculates $\mathscr{C}_{kp_i }^{i}/(1-\mathscr{C}_{kp_i}^{i})$ for all human predictions $p_i$ and classes $k \in \mathcal{C}(x)$, which requires $\mathcal{O}(hc)$ steps. Then it calculates the product term $\prod_{i \in \mathcal{H}} \mathscr{C}_{kp_i }^{i}/(1-\mathscr{C}_{kp_i}^{i})$ but only for each $k \in \mathcal{C}$ and when $p_i \in \mathcal{C}$. It chooses the pseudo label that maximizes this product term as $y^*$. Calculation of the product term again takes $\mathcal{O}(hc)$ steps and pseudo label calculation needs $\mathcal{O}(c)$ steps.  Finally, given $y^*$, it selects human subsets that satisfies $\mathscr{C}_{y^*p_i }^{i}/(1-\mathscr{C}_{y^*p_i}^{i}) >1$, which needs $\mathcal{O}(h)$ steps. Thus, the overall time complexity is $\mathcal{O}(hc)$, indicating that the algorithm's performance scales linearly with both the number of humans $h$ and the size of the conformal set $c$.

\begin{algorithm}
    \caption{Greedy Selection of Humans based on Conformal Sets $\mathcal{C}(x)$}
    \label{alg:greedy_selection}
    \begin{algorithmic}[1] 
        \Require $h, n, c \in \mathbb{N}$
        \Require conformal set of indices $\mathcal{C}(x)$
        \Require $n \times n$ matrices $\mathscr{C}^{i}$, $1 \leq i \leq h$
        \Require initial predictions $\hat{p}_i(x)$, $1 \leq i \leq h$
        
        \For{$i = 1$ to $h$}
            \For{$k = 1$ to $c$}
                \State $\mathcal{A} \gets \mathscr{C}^i_{\mathcal{C}(x)_k \hat{p}_i(x)}$  
                \State $\mathcal{B} \gets 1 - \mathscr{C}^i_{\mathcal{C}(x)_k \hat{p}_i(x)}$  
                \State $F[i][k] \gets \frac{\mathcal{A}}{\mathcal{B}}$  
            \EndFor
        \EndFor
        
        \For{$k = 1$ to $c$}
            \State $S[k] \gets 1$  
            \For{$i = 1$ to $h$}
                \If{$F[i][k] > 1$}
                    \State $S[k] \gets S[k] \times F[i][k]$  
                \EndIf
            \EndFor
        \EndFor
        
        \State $k^* \gets \arg\max_{1 \leq k \leq c} S[k]$  
        \State $y^* \gets \mathcal{C}(x)_{k^*}$  
        \State $S^* \gets \{1 \leq i \leq h \: | \: F[i][k^*] > 1\}$ 
    \end{algorithmic}
\end{algorithm}


        
        

\textbf{Remarks}. Note that to maximize the lower bound, we can set the user-specified tolerance level $\alpha \approx 0$. However, this may result in prediction sets that cover all classes, rendering them useless and uninformative for practical decision-making. That being noted, in our context—where a classifier is employed in a human-AI collaborative environment—we assume that the classifier chosen for such high-risk tasks has high predictive accuracy and is generally confident in its predictions. Moreover, we assume that the calibration data are well distributed across the classes. In the simulation study where we use accurate pre-trained models and well-distributed calibration data, we achieve a low average set size even though we set a low tolerance level $\alpha$, which results in nearly 100\% coverage. 

\textbf{Naive subset selection methods}. Several naive approaches to human subset selection exist. One such method involves setting $\mathcal{S}(x) = \mathcal{H}$, meaning all humans are employed for each data sample, followed by a combination method $\pi$ that applies a majority decision rule\footnote{Note that we chose the combination method $\pi$ to be the majority decision rule for simplicity. For future work, other combination frameworks can be experimented with.}. In our simulation study, we refer to this approach as "ALL HUMANS." Another method involves selecting a random subset $\mathcal{S}(x)$ from the set of human experts $\mathcal{H}$, ensuring that the average size of these human subsets aligns closely with the average subset size derived from our proposed greedy approach. This approach also utilizes a majority decision rule and is labeled "RANDOM SUBSET" in our simulations. Additionally, we employ a greedy selection method as described in Algorithm \ref{alg:greedy_selection}, which focuses on the top-$k$ prediction sets for each instance. The top-$k$ classes are determined by the output scores of the pre-trained classifier $\hat{f}$. Instead of using conformal sets $\mathcal{C}$ for human subset selection, this method relies on the top-$k$ prediction sets, while still applying the majority decision rule as the combination method $\pi$. In our simulations, we experiment with various values of $k$ and refer to this approach as "TOP-$k$." It is important to note that for both "ALL HUMANS" and "RANDOM SUBSET," the selected humans are required to make their final predictions based on conformal sets. In contrast, for "TOP-$k$," the selected humans must derive their final predictions from the top-$k$ prediction sets rather than from conformal sets.

\section{Experiments on CIFAR-10H Dataset}

In this section, we perform experiments using a dataset of natural images with real expert predictions for a multiclass classification task. We utilize several accurate deep neural network classifiers $\hat{f}$ to compute conformal scores for generating conformal sets, following methodologies similar to those in prior research \cite{Straitouri2022ImprovingEP}. Our proposed algorithm is benchmarked against naive subset selection methods and previous human-AI combination approaches that consider both single and multiple experts. Additionally, we compare our results with top-$k$ set-valued predictor baselines. We have included the detailed significance values for each experiment in Appendix \ref{appendix:sigval_cifar}.

This simulation study utilizes the (estimated) confusion matrix derived from real expert predictions for the multiclass classification task, along with the multinomial logit model defined in Eq. \ref{eq:multinomial_logit}, to evaluate our system's performance—consistent with findings from previous studies \cite{Straitouri2022ImprovingEP}. After selecting the subset of humans, we apply a combination method $\pi$, which follows a majority decision rule to determine the final outcome.

\subsection{Experimental Setup}

We conduct experiments on the CIFAR-10H dataset \cite{Peterson2019HumanUM}, which comprises 10,000 natural images sourced from the CIFAR-10 test set \cite{Krizhevsky2009LearningML}. Each image belongs to one of $ n=10 $ classes and includes approximately 50 expert predictions. For this dataset, we utilize three widely recognized deep learning models: DenseNet \cite{Huang2016DenselyCC}, ResNet \cite{He2015DeepRL}, and PreResNet \cite{He2016IdentityMI}, as similarly employed in \cite{Straitouri2022ImprovingEP}. We split the dataset into three subsets: calibration, estimation, and test set. In our classification tasks, we measure test data accuracy (empirical success probability). Additionally, in frameworks using conformal sets\footnote{In the simulation study, the tolerance levels $ \alpha $ are set at 0.10\%, 0.07\%, 0.05\%, 0.04\%, and 0.03\% for calibration data sizes $ l $ of 1,000, 1,500, 2,000, 2,500, and 3,000, respectively.}, we calculate average set sizes.

\subsection{Comparison with Naive Approaches}

\begin{figure*}[htbp!]
    \centering
    \begin{minipage}{0.43\textwidth}
        \includegraphics[width=8cm]{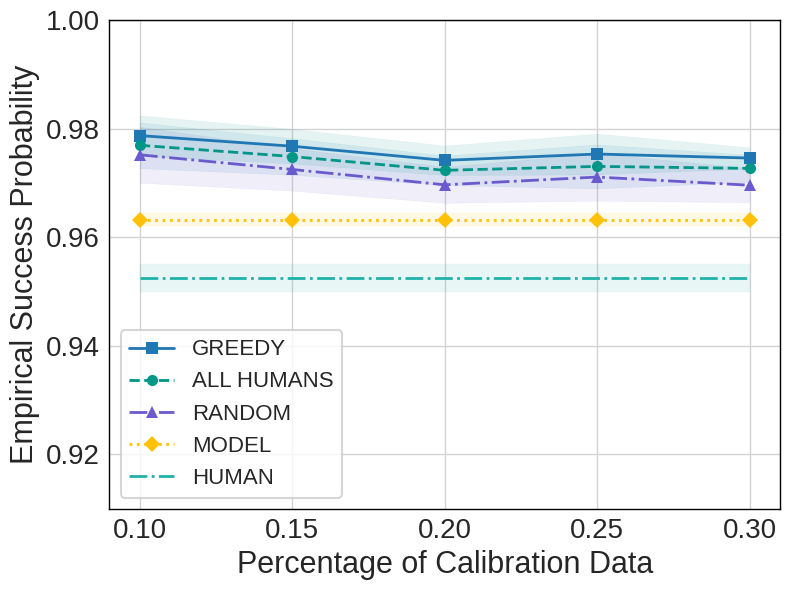} 
    
    \end{minipage}
    \hfill
    \begin{minipage}{0.467\textwidth}
        \includegraphics[width=8cm]{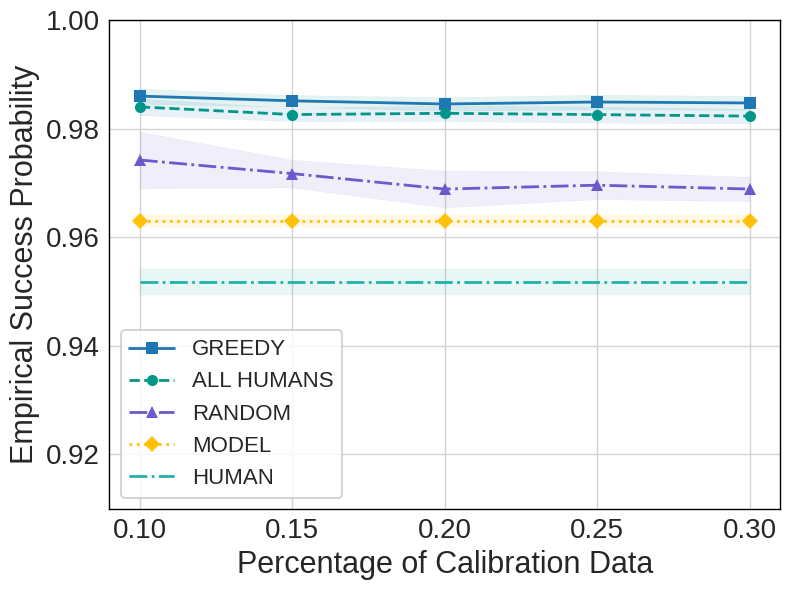} 
        
    \end{minipage}

    \caption{Empirical success probability of a multiple expert system utilizing the proposed greedy human subset selection algorithm in comparison with naive approaches across varying percentages of calibration data on the CIFAR-10H dataset. The estimation data matches the calibration data proportionally. The left figure presents results for 3 experts, while the right shows results for 5 experts. "ALL HUMANS" and "RANDOM" refer to all-human and random subset selection methods, respectively, applying a majority decision rule. "MODEL" and "HUMAN" indicate the performance of the classifier and a single expert. The classifier employed is Densenet. The average sizes of the conformal sets are 3.75, 4.79, 5.66, 5.69, and 5.92 for calibration percentages of 10\%, 15\%, 20\%, 25\%, and 30\%, respectively. Results are averaged over 10 runs, with shaded regions representing one standard deviation from the mean.}
    \label{fig:cifar_naive}
    \Description{Line plot showing accuracy of the proposed approach and other naive methods on the CIFAR10H dataset. Shaded regions refer to one standard deviation away from the mean.}
\end{figure*}

In Figure \ref{fig:cifar_naive}, we compare the empirical success probability $\mathbb{P}[\hat{Y} = Y; \mathcal{C}|H]$ on the test data for our proposed greedy selection algorithm with naive methods, such as randomly choosing subsets of humans of average size $\tau$ (the average subset size of our greedy selection approach) and selecting all humans followed by a majority decision rule. We also visualize the results for the best single human performance and the pre-trained model. While the all-human approach reflects the intuition to favor majority decisions, leading to high empirical success probabilities, it doesn't always yield the best outcomes. In contrast, our greedy algorithm based on the conformal set demonstrates superior effectiveness in selecting human subsets.

\subsection{Comparison with Top-$k$ Set-Valued Predictors}

\begin{figure}[htp!] 
    \centering
    \includegraphics[width=0.43\textwidth]{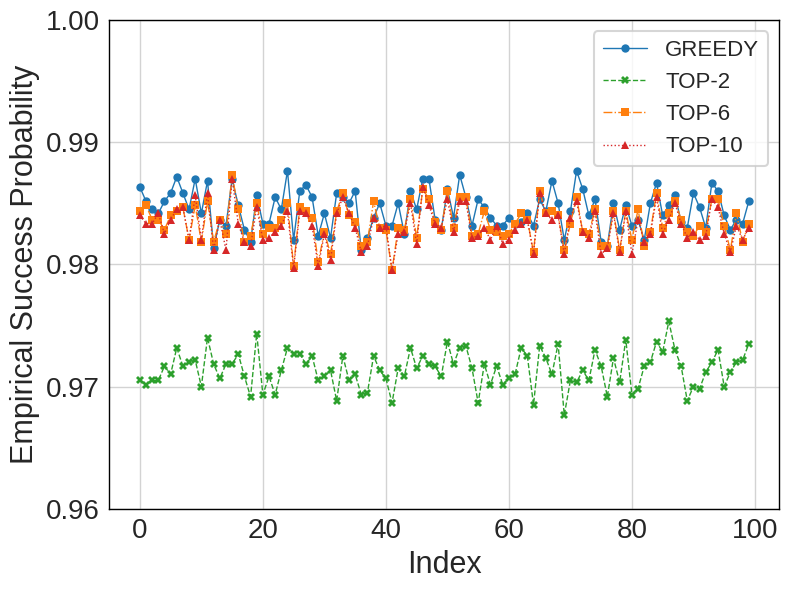} 
    \caption{Empirical success probability of our proposed greedy selection algorithm in comparison with top-$k$ set predictor framework for different values of $k$ on the CIFAR-10H dataset using the Resnet classifier for 100 runs (using $h=5$). The empirical average set size of the conformal sets is 5.2718. The calibration and estimation sets each have a size of 2,000.} 
    \label{fig:cifar10h_set_predictors_results} 
    \Description{Line plot for 100 runs showing accuracy of the proposed approach and other top-k methods.}
\end{figure}

Figure \ref{fig:cifar10h_set_predictors_results} compares our conformal multi-expert method, which uses a greedy algorithm for selecting human subsets, with a top-$k$ set predictor framework that also employs a greedy selection approach based on the top-$k$ sets. Our method shows a higher empirical success probability.

\subsection{Comparison with Human-AI Combination Approaches}
In Table \ref{tab:cifar10h_results_baselines}, we compare the empirical success probability of our conformal-based greedy algorithm with existing methods that integrate pre-trained classifier outputs and human predictions. The findings reveal two key insights: multi-expert collaboration yields a higher success probability than relying on a single expert, and our greedy algorithm effectively selects human subsets for classification. Notably, even with a limited set of options, our approach surpasses previous baselines that allowed full label access for humans, demonstrating the conformal predictor's ability to identify meaningful classes for each instance.

\begin{table}[t!]
    \setlength{\tabcolsep}{0.75pt}
    \small  
    \centering
    \caption{Empirical success probability (in \%) achieved by our conformal multi-expert approach during test using the greedy selection algorithm using three accurate neural network-based pre-trained classifiers and five multiple experts ($h=5$) in comparison with previous baselines on the CIFAR-10H dataset. The single expert's empirical success probability at solving the (original) multiclass task is $\approx 95.24\%$. The calibration and estimation sets each have a size of 1,000. The values are averaged over 10 runs.}
    \begin{tabular}{lccc} 
    \toprule
        MODEL NAME & DENSENET &	RESNET & PRERESNET \\ 
        \midrule 
        Pre-trained Model Alone & 96.29 $\pm$
0.09 & 92.70 $\pm$ 0.20 & 94.41 $\pm$
0.12 \\ 
        \multicolumn{1}{l}{SINGLE EXPERT APPROACHES} &  \multicolumn{3}{c}{}   \\ 
        \citet{Kerrigan2021CombiningHP} (MAP temp. scal.)& 97.81 $\pm$ 0.25 & 97.22
$\pm$ 0.40 & 97.56 $\pm$ 0.36 \\ 
        \citet{Straitouri2022ImprovingEP}  & 97.70
        $\pm$ 0.09 & 96.63 $\pm$ 0.13 & 97.05 $\pm$ 0.18 \\ 
        \multicolumn{1}{l}{MULTIPLE EXPERT APPROACHES} &  \multicolumn{3}{c}{}  \\
        \citet{Singh2023OnSS} (Greedy algorithm) & 95.15 $\pm$ 0.22 & 95.18 $\pm$
0.21 & 95.25 $\pm$ 0.20 \\ 
        \citet{Singh2023OnSS} (Mode approach) & 97.93 $\pm$ 0.13 & 97.23 $\pm$
0.19 & 97.48 $\pm$ 0.16 \\ 
        Proposed Greedy Selection Method& \textbf{98.48 $\pm$ 0.22} & \textbf{98.48 $\pm$ 0.15} & \textbf{98.10 $\pm$ 0.38} \\ 
\bottomrule
    \end{tabular}

     \label{tab:cifar10h_results_baselines}
\end{table}

\section{Experiments on ImageNet-16H Dataset}

In this section, we conduct experiments using a different dataset of natural images that includes real expert predictions and features a greater number of classes. For the pretrained models $\hat{f}$, we utilize the VGG19 deep neural network classifier, as provided by Steyvers \etal \cite{Steyvers2022BayesianMO}, which has been fine-tuned over 10 epochs. We assess the performance of our proposed greedy subset selection algorithm by comparing it to naive methods for human subset selection. Additionally, we evaluate our approach against top-$k$ set-valued predictors that implement a modified greedy strategy for selecting human subsets based on the top-$k$ set. Detailed significance values for all experiments can be found in Appendix \ref{appendix:sigval_imagenet}. 

\subsection{Experimental Setup}

We experiment with the ImageNet-16H dataset \cite{Steyvers2022BayesianMO}, which comprises 1,200 unique images derived from a subset of the ImageNet Large Scale Visual Recognition Challenge (ILSVRC) 2012 database \cite{Russakovsky2014ImageNetLS}. This dataset includes approximately six predictions made by human experts for each image. Each image can be classified into one of the $ n = 16 $ categories based on separate human annotations. To increase the difficulty of the annotations for both humans and classifiers, the images are distorted using spatial frequency phase noise with a frequency of $\omega = 80$ Additionally, we randomly divided the images into three subsets: the calibration set, estimation set, and the test set. Typically, these subsets contain 240 images each for the calibration sets and estimation sets, and 720 images for the test set, unless stated otherwise in the experimental results. We use the calibration set\footnote{The tolerance levels $\alpha$ are set at 1.10\%, 0.83\%, 0.66\%, and 0.55\% for calibration data of sizes 180, 240, 300, and 360, respectively.} to compute the conformal scores necessary for the conformal set predictor. Then we utilize the estimation set (together with the calibration set) to estimate the human confusion matrix. Finally, we utilize the test set to assess the empirical success probability of the experts as applied in various approaches. To illustrate the reduced size of the conformal sets, we also compute the empirical average sizes of these sets.

\subsection{Comparison with Naive Approaches}

\begin{figure}[htp!] 
    \centering
    \includegraphics[width=0.43\textwidth]{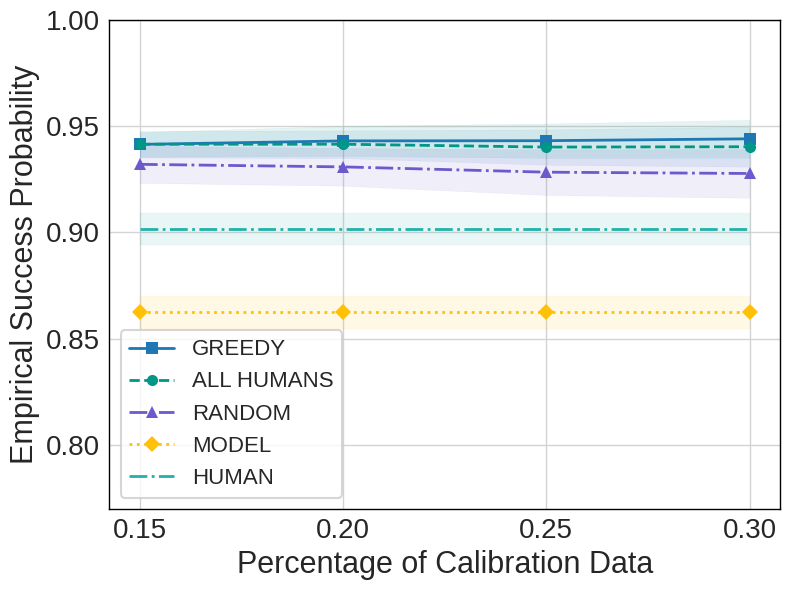} 
    \caption{Empirical success probability of the multi-expert system using the proposed greedy human subset selection algorithm compared to naive approaches across various calibration data percentages on the ImageNet-16H dataset, with results displayed for 5 experts. "ALL HUMANS" and "RANDOM" refer to all-human and random subset selection methods, while "MODEL" and "HUMAN" indicate the performance of the classifier and a single expert, respectively. The estimation data is proportionally matched to the calibration data. The average sizes of the conformal sets are 2.67, 2.82, 3.26, and 3.39 for calibration percentages of 15\%, 20\%, 25\%, and 30\%. Results are averaged over 100 runs, with shaded regions representing one standard deviation from the mean.}
    \label{fig:imagenet_naive}
    \Description{Line plot showing accuracy of the proposed approach and other naive methods on the Imagenet16H dataset. Shaded regions refer to one standard deviation away from the mean.}
\end{figure}

Figure \ref{fig:imagenet_naive} illustrates the comparison of empirical success probabilities among the proposed greedy selection method, the random subset selection method, and the all-humans approach when the combination method $\pi$ is set to the majority decision rule. Observably, the proposed greedy method demonstrates superior performance compared to the naive approaches.

\subsection{Comparison with Top-$k$ Predictors}

\begin{figure}[htp!] 
    \centering
    \includegraphics[width=0.43\textwidth]{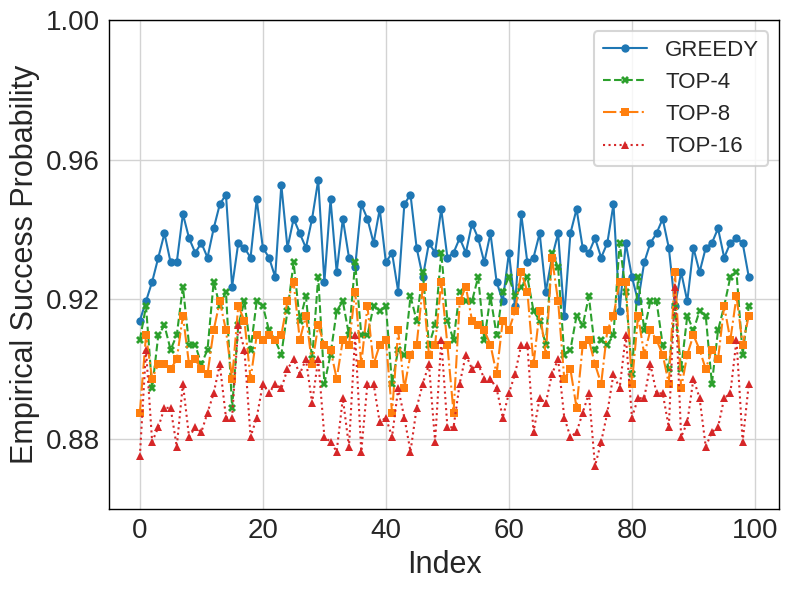} 
    \caption{Empirical success probability achieved by the multi-expert system utilizing the proposed greedy human subset selection algorithm in comparison with top-$k$ set-valued predictors on the ImageNet-16H dataset over 100 runs. The figure illustrates results with 3 experts. Both the calibration and estimation sets comprise 240 images each. The empirical average size of the conformal sets is 2.8247.}
    \label{fig:set_baselines_results_imagenet_1}
    \Description{Line plot showing accuracy of the proposed approach and topk approaches for 100 runs.}
\end{figure}

\begin{figure}[htp!] 
    \centering
    \includegraphics[width=0.43\textwidth]{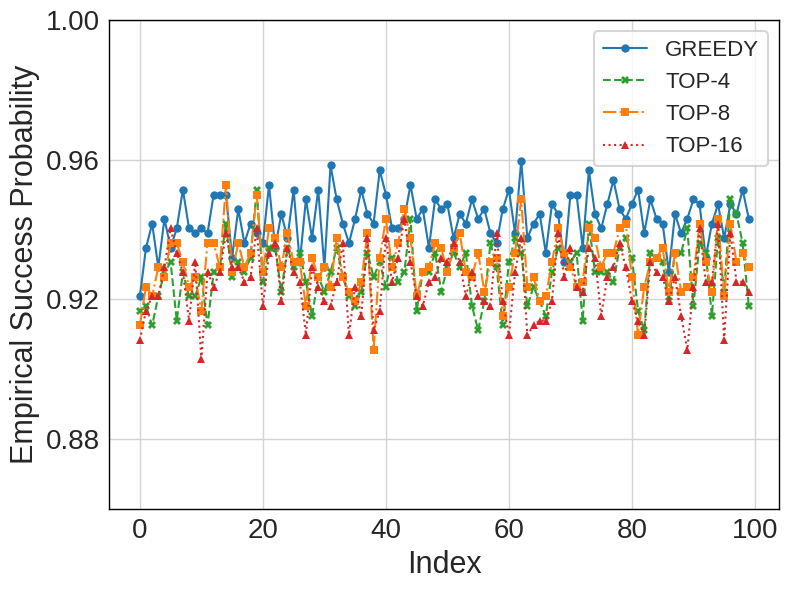} 
    \caption{Empirical success probability of the system using the proposed greedy human subset selection algorithm compared to top-$k$ set-valued predictors on the ImageNet-16H dataset over 100 runs. The figure presents results with 5 experts and maintains the same settings as Figure \ref{fig:set_baselines_results_imagenet_1}.}
    \label{fig:set_baselines_results_imagenet_2}
    \Description{Line plot showing accuracy of the proposed approach and topk approaches for 100 runs.}
\end{figure}

Figures \ref{fig:set_baselines_results_imagenet_1} and \ref{fig:set_baselines_results_imagenet_2} compare our conformal multi-expert approach using the greedy algorithm to select human subsets against the framework employing top-$k$ set predictors using mode approach for subset selection. Our framework demonstrates superior empirical success probability, outperforming both single-expert and multi-expert baselines, even when the average prediction set size of the conformal sets is smaller than $k$.

\subsection{What happens as the number of multiple experts $h$ increases?}
\begin{figure}[htp!] 
    \centering
    \includegraphics[width=0.43\textwidth]{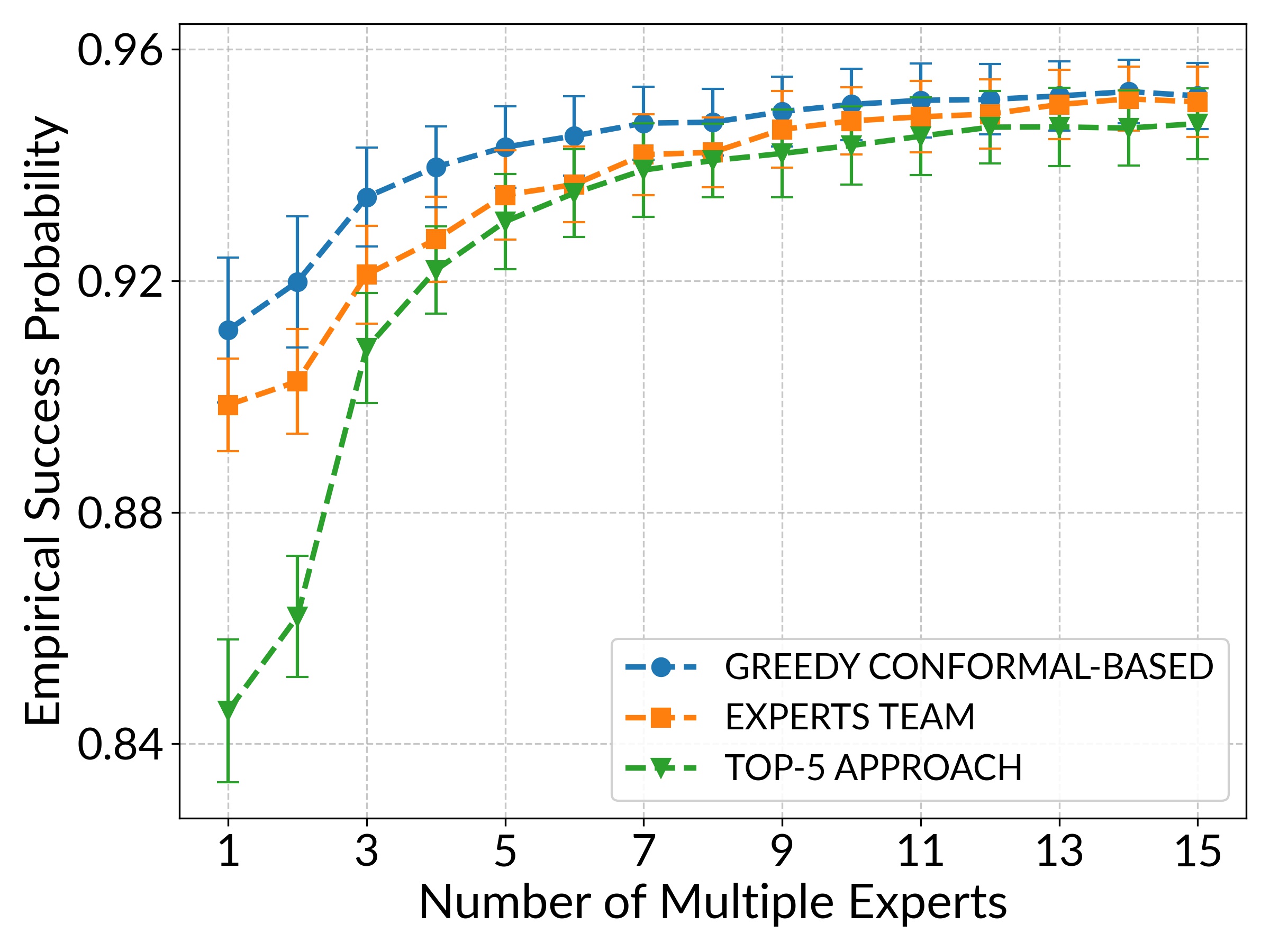} 
    \caption{Empirical success probability of our proposed greedy selection algorithm in comparison with the expert team and top-$5$ set predictor framework for different number of experts $h$ on the ImageNet-16H dataset. The values are averaged over 100 runs. The empirical average set size of the conformal sets for all values of $h$ is 2.8247. The size of the calibration and estimation sets is $l=240$. The error bars indicate one standard deviation from the mean.} 
    \label{fig:imagenet_varying_m} 
    \Description{Line plot showing accuracy of the proposed approach, expert team and top8 approach for number of experts from 1 to 15 on the ImageNet16H dataset. Shaded regions refer to one standard deviation away from the mean.}
\end{figure}

In Figure \ref{fig:imagenet_varying_m}, we compare the empirical success probabilities of our multi-expert system using the greedy subset selection method against a human-only expert team and a top-$k$ predictor with a top-$k$ subset selection algorithm (where $k=5$). The expert team reflects majority predictions without relying on prediction sets or models. Our results show that as the number of human experts increases, our conformal set-based greedy selection approach outperforms both the expert team and the top-$5$ methods, remaining effective even with a large pool of human experts.

\section{Discussion}

In this section, we explore the assumptions and limitations of the proposed approach.

\textbf{Framework.} The proposed greedy algorithm relies on an estimated confusion matrix for human performance, based on real expert predictions and following a straightforward Maximum Likelihood Estimation approach similar to that of Kerrigan et al. \cite{Kerrigan2021CombiningHP}. Exploring more sophisticated methods for estimating the confusion matrix would be advantageous. Additionally, the assumption of independence among experts may not hold, as their decisions can influence one another. In our framework, we set the tolerance level $\alpha$ close to zero, which minimally impacts average conformal set sizes, but this may not always be practical. Future research should examine scenarios where a clear trade-off exists between the tolerance level $\alpha$ and average conformal set sizes.

\textbf{Empirical results.} The experimental results show that the proposed greedy selection of human subsets is non-trivial, highlighting its advantages over majority decisions or random expert choices. However, our experiments assume well-distributed calibration data, indicating a need for further exploration in scenarios with multiple classes and class imbalances. Additionally, we treat all experts as equally important, though some may offer greater value. Future research should examine settings where experts have varying levels of expertise and assess the likelihood of including high-importance experts in the selected human subset.

\textbf{Broader impact.} The greedy subset selection of human experts enables a realistic human-AI collaboration, where each expert chooses from a refined set of options. Future work could enhance this framework by accurately modeling experts' class preferences and interdependencies, as well as exploring alternative collaboration methods, such as a scoring system instead of simple subset selection.

\section{Conclusion}
\label{conclusion}

We have looked at the challenge of human-AI collaboration among multiple experts through the framework of subset selection. Our theoretical analysis demonstrates the conditions under which selecting multiple experts from conformal sets is more advantageous than choosing from the entire label space. Inspired by these findings, we propose a greedy algorithm for selecting human subsets based on conformal sets to improve classification performance during inference. We have shown that this method outperforms both naive subset selection approaches and greedy strategies based on top-$k$ prediction sets.

\begin{acks}
Guohao Shen’s research is (partially) supported by the Hong Kong Research Grants Council (Grant No. 15305523) and research grants from The Hong Kong Polytechnic University (Grant No. P0048718 and No. P0041243).
\end{acks}


\bibliographystyle{ACM-Reference-Format} 
\balance
\bibliography{sample}

\clearpage
\onecolumn
\appendix

\section{Proofs}

\subsection{Proof of Lemma 1}
\label{appendix_proofs_1}

LEMMA 1. Given $h$ human expert predictions $H=[p_1, p_2, \ldots, p_h]$, conformal set $\mathcal{C}$ with tolerance level $\alpha$, and the $i$th expert’s conditional success probability $\mathscr{C}^i$, the lower bound on the accuracy of the combined framework is given as

\begin{equation}
\mathbb{E}[ \indicator( \{\hat{Y}=y  \} \: \cap \: \{y \in \mathcal{C}\})] \geq \mathbb{P} [ \: \prod\limits _{i \in [h]} \dfrac{\mathscr{C}_{y p_i }^{i}}{1-\mathscr{C}_{y p_i}^{i}} > 1 ] \cdot (1-\alpha)
\label{eq:expectation_bound}
\end{equation}

PROOF:

\begin{align*}
    \mathbb{E}[\indicator ( \{\hat{Y}=y \} \: \cap \: \{y \in \mathcal{C} \})] &= \mathbb{P}[\hat{Y}=y| H, y \in \mathcal{C}] \cdot \mathbb{P}[y \in C|H] \\ 
    &\geq \mathbb{P}[\hat{Y}=y|H,y \in \mathcal{C}] \cdot (1-\alpha)\\
    &= \mathbb{P} [ \: y = \argmax\limits_{k \in \mathcal{C}} \:\:\: \prod\limits_{i \in [h]} \mathscr{C}_{kp_i}^{i} \: ] \cdot (1-\alpha)\\
    &= \mathbb{P} [ \: \prod\limits_{i \in [h]} \mathscr{C}_{yp_i} ^{i} > \max\limits_{k \neq y, k \in \mathcal{C}} \:\:\:   \prod\limits_{i \in [h]} \mathscr{C}_{k p_i}^{i} \: ] \cdot (1-\alpha)\\
    &\geq \mathbb{P}[ \: \prod\limits_{i \in [h]} \mathscr{C}_{yp_i} ^{i} >   \prod\limits_{i \in [h]} \max\limits_{k \neq y, k \in \mathcal{C}}\:\:\: \mathscr{C}_{k p_i}^{i} \: ] \cdot (1-\alpha)\\
    &\geq \mathbb{P} [  \: \prod\limits_{i \in [h]} \mathscr{C}_{y p_i} ^{i} >  \prod\limits_{i \in [h]} (1-\mathscr{C}_{yp_i}^{i}) \: ] \cdot (1-\alpha)\\
    &= \mathbb{P} [ \: \prod\limits _{i \in [h]} \dfrac{\mathscr{C}_{y p_i }^{i}}{1-\mathscr{C}_{y p_i}^{i}} > 1  \: ] \cdot (1-\alpha)
\end{align*}








\subsection{Proof of Lemma 2}
\label{appendix_proofs_2}

LEMMA 2. Let $\epsilon = \prod\limits _{i \in [h]} \dfrac{\mathscr{C}_{yp_i}^{i}}{1-\mathscr{C}_{yp_i}^{i}} - \prod\limits _{i \in [h]} \dfrac{C_{yp_i}^{i}}{1-C_{yp_i }^{i}}$. Assuming $\epsilon > 1$ and $\alpha=0$, the combination framework of multiple human expert predictions that utilizes a conformal set achieves a tighter lower bound on accuracy,

\begin{equation}
\mathbb{P} [ \prod\limits _{i \in [h]} \dfrac{C_{yp_i}^{i}}{1-C_{y p_i}^{i}} > \dfrac{1-\hat{f}_{y}}{\hat{f}_{y}} ] \leq \mathbb{P} [ \: \prod\limits _{i \in [h]} \dfrac{\mathscr{C}_{yp_i}^{i}}{1-\mathscr{C}_{yp_i}^{i}} > 1 ].
\end{equation}

PROOF: To prove this result, we start from the right side of the inequality.

\begin{align*}
\mathbb{P} [ \: \prod\limits _{i \in [h]} \dfrac{\mathscr{C}_{yp_i }^{i}}{1-\mathscr{C}_{yp_i}^{i}} > 1 ] &= \mathbb{P} [ \: \prod\limits _{i \in [h]} \dfrac{\mathscr{C}_{yp_i}^{i}}{1-\mathscr{C}_{yp_i}^{i}} -1 > 0 ]\\
&= \mathbb{P} [ \: \prod\limits _{i \in [h]} \dfrac{\mathscr{C}_{yp_i}^{i}}{1-\mathscr{C}_{yp_i}^{i}} -1 + \prod\limits _{i \in [h]} \dfrac{C_{yp_i}^{i}}{1-C_{yp_i}^{i}} - \prod\limits _{i \in [h]} \dfrac{C_{yp_i}^{i}}{1-C_{yp_i}^{i}} > 0 ]\\
&= \mathbb{P} [ \:  \prod\limits _{i \in [h]} \dfrac{C_{yp_i}^{i}}{1-C_{yp_i}^{i}} +\prod\limits _{i \in [h]} \dfrac{\mathscr{C}_{yp_i}^{i}}{1-\mathscr{C}_{yp_i}^{i}} - \prod\limits _{i \in [h]} \dfrac{C_{yp_i}^{i}}{1-C_{yp_i}^{i}} -1  > 0 ]\\
&= \mathbb{P} [ \:  \prod\limits _{i \in [h]} \dfrac{C_{yp_i}^{i}}{1-C_{yp_i}^{i}} + \epsilon -1  > 0 ]\\
&\geq \mathbb{P} [ \:  \prod\limits _{i \in [h]} \dfrac{C_{yp_i}^{i}}{1-C_{yp_i}^{i}} > 0 ]\text{ since we assume }\epsilon > 1\\
&\geq \mathbb{P} [ \prod\limits _{i \in [h]} \dfrac{C_{yp_i}^{i}}{1-C_{yp_i}^{i}} > \dfrac{1-\hat{f}_{y}}{\hat{f}_{y}} ]
\end{align*}

\section{Significance Values for the Experiments}

\subsection{CIFAR-10H Dataset}
\label{appendix:sigval_cifar}

We show significance values for the accuracy differences in our experiments, comparing our greedy method to ALL HUMANS method using the CIFAR-10H dataset. We used either paired t-test or Wilcoxon signed-rank test (based on data normality). 

\begin{table}[h!]
    \small  
    \centering
    \caption{Significance values for the CIFAR-10H experiments with 3 multiple experts comparing the accuracy of our proposed method with the ALL HUMANS approach in Figure \ref{fig:cifar_naive}.}
    \begin{tabular}{ccc} 
        \toprule
        Calib. data (in \%) &  p-value & Conclusion \\ 
        \midrule 
        10  & 0.002 & Our method is statistically significantly better     \\
        15  & 0.002 & Our method is statistically significantly better     \\
        20	& 0.004 & Our method is statistically significantly better     \\
        25	& 0.002 & Our method is statistically significantly better     \\
        30	& 0.002 & Our method is statistically significantly better    \\
        \bottomrule
    \end{tabular}
    \label{tab:sigval_cifar_naive_3}
\end{table}

\begin{table}[h!]
    \small  
    \centering
    \caption{Significance values for the CIFAR-10H experiments with 5 multiple experts comparing the accuracy of our proposed method with the ALL HUMANS approach in Figure \ref{fig:cifar_naive}.}
    \begin{tabular}{ccc} 
        \toprule
        Calib. data (in \%) &  p-value & Conclusion \\ 
        \midrule 
        10  & 0.002 & Our method is statistically significantly better     \\
        15  & 0.002 & Our method is statistically significantly better     \\
        20	& 0.008 & Our method is statistically significantly better     \\
        25	& 0.002 & Our method is statistically significantly better     \\
        30	& <0.001 & Our method is statistically significantly better    \\
        \bottomrule
    \end{tabular}

     \label{tab:sigval_cifar_naive_5}
\end{table}

In the table that follows, we show significance values for the differences in accuracy in our CIFAR-10H experiments, comparing our proposed greedy method to the top-\textit{6} and top-\textit{10} approaches. We used either paired t-test or Wilcoxon signed-rank test (depending on data normality). 

\begin{table}[h!]
    \small  
    \centering
    \caption{Significance values for the CIFAR-10H experiments with 5 multiple experts comparing the accuracy of our proposed method with the top-\textit{6} and top-\textit{10} approach in Figure \ref{fig:cifar10h_set_predictors_results}.}
    \begin{tabular}{ccc} 
        \toprule
         &  p-value & Conclusion \\ 
        \midrule 
        ours vs top-6  & <0.001  & Our method is statistically significantly better    \\
        ours vs top-10  & <0.001  & Our method is statistically significantly better    \\
        \bottomrule
    \end{tabular}

     \label{tab:sigval_cifar_top}
\end{table}


\subsection{ImageNet-16H Dataset}
\label{appendix:sigval_imagenet}

We now show significance values for the accuracy differences in our experiments, comparing our greedy method to the ALL HUMANS method using the ImageNet-16H dataset. We used either paired t-test or Wilcoxon signed-rank test (based on data normality). We observe that with relatively small calibration data, our proposed approach is not statistically significantly better than the ALL HUMANS approach. This highlights the necessity of having sufficient calibration when using our proposed greedy method.

\begin{table}[h!]
    \small  
    \centering
    \caption{Significance values for the ImageNet-16H experiments with 5 multiple experts comparing the accuracy of our proposed method with the ALL HUMANS approach in Figure \ref{fig:imagenet_naive}.}
    \begin{tabular}{ccc} 
        \toprule
        Calib. data (in \%) &  p-value & Conclusion \\ 
        \midrule 
        15  & 0.926 & No significant difference     \\
        20	& <0.001 & Our method is statistically significantly better     \\
        25	& <0.001 & Our method is statistically significantly better     \\
        30	& <0.001 & Our method is statistically significantly better    \\
        \bottomrule
    \end{tabular}

     \label{tab:sigval_imagenet_naive_5}
\end{table}

In the following tables, we show significance values for the differences in accuracy in our ImageNet-16H experiments, comparing our proposed greedy method to the top-\textit{4}, top-\textit{8}, and top-\textit{16} approaches. We used either paired t-test or Wilcoxon signed-rank test (depending on data normality). 

\begin{table}[H]
    \small  
    \centering
    \caption{Significance values for the ImageNet-16H experiments with 3 multiple experts comparing the accuracy of our proposed method with the top-\textit{4}, top-\textit{8}, and top-\textit{16} approach in Figure \ref{fig:set_baselines_results_imagenet_1}.}
    \begin{tabular}{ccc} 
        \toprule
         &  p-value & Conclusion \\ 
        \midrule 
        ours vs top-4  & <0.001  & Our method is statistically significantly better    \\
        ours vs top-8  & <0.001  & Our method is statistically significantly better    \\
        ours vs top-16  & <0.001  & Our method is statistically significantly better    \\
        \bottomrule
    \end{tabular}

     \label{tab:sigval_imagenet_top_3}
\end{table}

\begin{table}[H]
    \small  
    \centering
    \caption{Significance values for the ImageNet-16H experiments with 5 multiple experts comparing the accuracy of our proposed method with the top-\textit{4}, top-\textit{8}, and top-\textit{16} approach in Figure \ref{fig:set_baselines_results_imagenet_1}.}
    \begin{tabular}{ccc} 
        \toprule
         &  p-value & Conclusion \\ 
        \midrule 
        ours vs top-4  & <0.001  & Our method is statistically significantly better    \\
        ours vs top-8  & <0.001  & Our method is statistically significantly better    \\
        ours vs top-16  & <0.001  & Our method is statistically significantly better    \\
        \bottomrule
    \end{tabular}
     \label{tab:sigval_imagenet_top_5}
\end{table}

\section{Implementation Details}
\label{appendix_imp_details}

To conduct our experiments, we utilize PyTorch 2.1.0, NumPy 1.26.4, and Scikit-learn 1.0.2 in Python 3.9.19. To ensure reproducibility, we kept a fixed random seed across all random processes for each run. 

\textbf{CIFAR-10H.}
For each image in this dataset, there are approximately 50 expert predictions wherein each image can be categorized in one of the 10 classes.

\textbf{Imagenet-16H.}
This dataset comprises only 1200 samples, with a portion allocated for model training. The classes are chair, oven, knife, bottle, keyboard, clock, boat, bicycle, airplane, truck, car, elephant, bear, dog, cat, and bird. The original labels from the ILSRVR database are used as ground truth labels. To compute the confusion matrix for the humans, we utilize the empirical expert distribution corresponding to the estimation and calibration data, which are not used for testing. We simulate multiple humans for testing by sampling from the expert distribution for test data. This distribution represents a categorical distribution for each class provided by the human experts.


\textbf{Execution Time and Memory Consumption} For our experiments, we evaluate our proposed approach on the CIFAR-10H and ImageNet-16H datasets using an NVIDIA A100 GPU accelerator. Our experimental setup utilizes CUDA version 11.8 to accelerate the computationally intensive workloads involved in training and evaluating our models on these datasets. For the CIFAR-10H dataset, each experimental run had modest memory requirements, consuming only a few megabytes, and completed swiftly within seconds. In contrast, the ImageNet-16H dataset posed a more computational challenge, requiring several minutes to complete.

\section{TESTING Different values of the tolerance level $\alpha$}

We show results for the proposed greedy algorithm for different $\alpha$ values in the CIFAR-10H dataset experiments, using 5 experts and a calibration proportion of 20\%. Setting $\alpha$ involves balancing desired coverage with implications on prediction set sizes. Our findings show that increasing $\alpha$ leads to a smaller average set size and a reduced empirical success probability. Since the reduction in the average set size is not drastic (an acceptable trade-off), it is reasonable to select a small $\alpha$ value in this setting.

\begin{table}[H]
    \small  
    \centering
    \caption{Mean set size and accuracy of the proposed greedy algorithm for different values of $\alpha$ in the CIFAR-10H dataset.}
    \begin{tabular}{ccc} 
        \toprule
        $\alpha$ (in \%) &  Mean set size & Accuracy (in \%) \\ 
        \midrule 
        1 &	1.17 &	98.48 \\
        3 &	1.01 & 	97.00 \\
        5 &	0.97 &	94.93 \\
        7 &	0.94 &	92.92  \\
        \bottomrule
    \end{tabular}
     \label{tab:changing_alpha}
\end{table}

\section{Additional Remarks on the Significance and Limitations of the Study}

The accuracy improvement of the proposed framework is crucial for safety-critical applications, such as classification in medical AI or autonomous driving, where tolerance to error is minimal. A limitation of our approach occurs when the calibration data is small. In CIFAR-10H data, when calibration data is 1\% (100 samples) for h=5, significance testing using paired t-test shows that our greedy framework and ALL HUMANS do not differ significantly in accuracy (p-value=0.09). In addition, we did not test our method beyond image classification due to the lack of publicly available datasets with multiple expert predictions per sample and more than 3 classes. Datasets like Hatespeech, COMPASS, and NIH Chest X-ray have real human predictions but are limited to 2 or 3 classes, restricting the analysis of the conformal predictor.


\end{document}